\documentclass[5p,times]{elsarticle}

\usepackage{lineno,hyperref}
\usepackage{graphicx}
\usepackage{float}
\usepackage{caption}
\usepackage{amssymb}
\usepackage{amsmath}
\usepackage{subcaption}
\usepackage{multirow}
\usepackage{booktabs}
\usepackage{threeparttable}
\usepackage[table,xcdraw]{xcolor}
\modulolinenumbers[5]
\journal{}

\begin{document}

\begin{frontmatter}

\title{ODformer: Spatial-Temporal Transformers
for Long Sequence  Origin-Destination Matrix Forecasting Against Cross Application Scenario }


\author[mymainaddress]{Jin Huang}
\ead{huangjin@m.scnu.edu.cn}

\author[mymainaddress]{Bosong Huang}
\ead{bosonghuang@m.scnu.edu.cn}

\author[mysecondaryaddress]{Weihao Yu}
\ead{yuwh3@chinatelecom.cn}

\author[mymainaddress]{Jing Xiao}
\ead{xiaojing@scnu.edu.cn}

\author[mymainaddress]{Ruzhong Xie}
\ead{rzxie@m.scnu.edu.cn}

\author[mysecondaryaddress]{Ke Ruan\corref{mycorrespondingauthor}}
\cortext[mycorrespondingauthor]{Corresponding author}
\ead{ruanke@chinatelecom.cn}

\address[mymainaddress]{South China Normal University, Guangzhou, China}
\address[mysecondaryaddress]{Research Institute of China Telecom Corporate Ltd., Guangzhou, China}

\begin{abstract}
Origin-Destination (OD) matrices record directional flow data between pairs of OD regions. The intricate spatiotemporal dependency in the matrices makes the OD matrix forecasting (ODMF) problem not only intractable but also non-trivial. However, most of the related methods are designed for very short sequence time series forecasting in specific application scenarios, which cannot meet the requirements of the variation in scenarios and forecasting length of practical applications. To address these issues, we propose a Transformer-like model named ODformer, with two salient characteristics: 
(i) the novel OD Attention mechanism, which captures special spatial dependencies between OD pairs of the same origin (destination), greatly improves the ability of the model to predict cross application scenarios after combining with 2D-GCN that captures spatial dependencies between OD regions. 
(ii) a PeriodSparse Self-attention that effectively forecasts long sequence OD matrix series while adapting to the periodic differences in different scenarios. 
Generous experiments in three application backgrounds (i.e., transportation traffic, IP backbone network traffic, crowd flow) show our method outperforms the state-of-the-art methods.
\end{abstract}


\end{frontmatter}

\section{Introduction}%
\begin{figure*}[tb]
    \centering
    \subfloat[NYC]{\label{1a}
            \centering
            \includegraphics[scale=0.25]{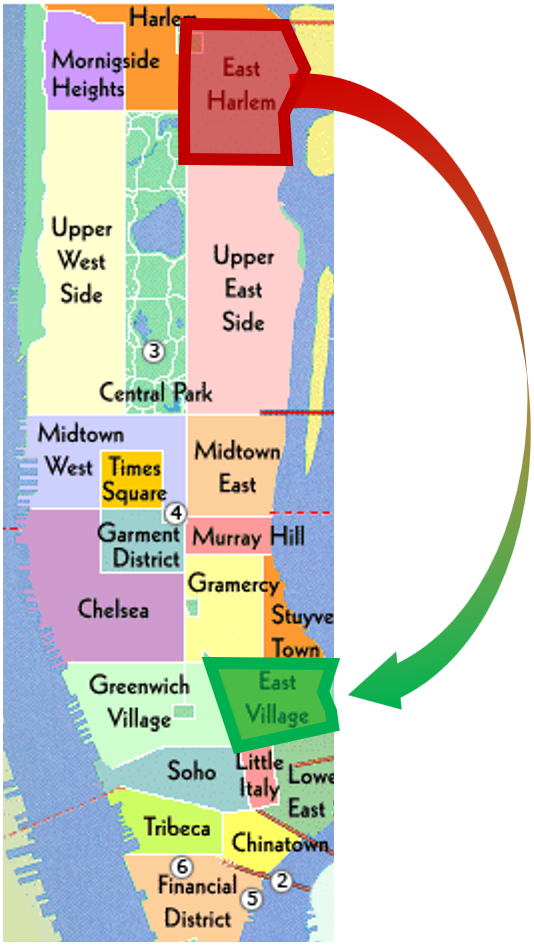}
            }
    \subfloat[JHT]{\label{1b}
            \centering
            \includegraphics[scale=0.25]{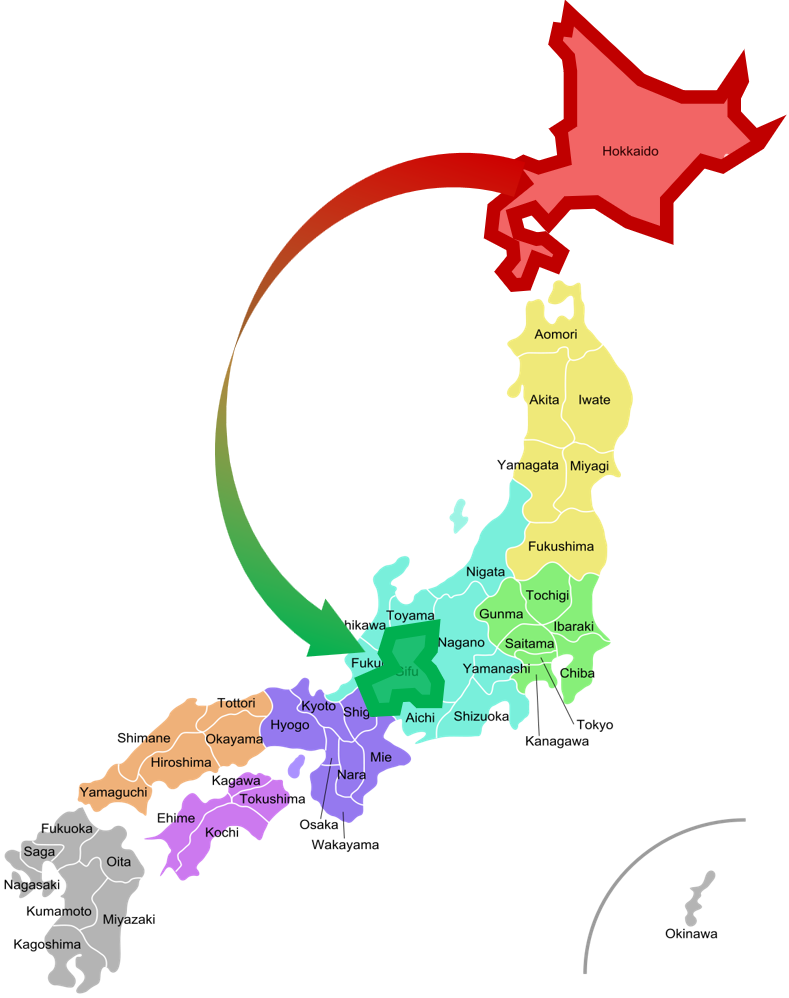}
            }%
    \subfloat[GÉANT ]{\label{1c}
            \centering
            \includegraphics[scale=0.22]{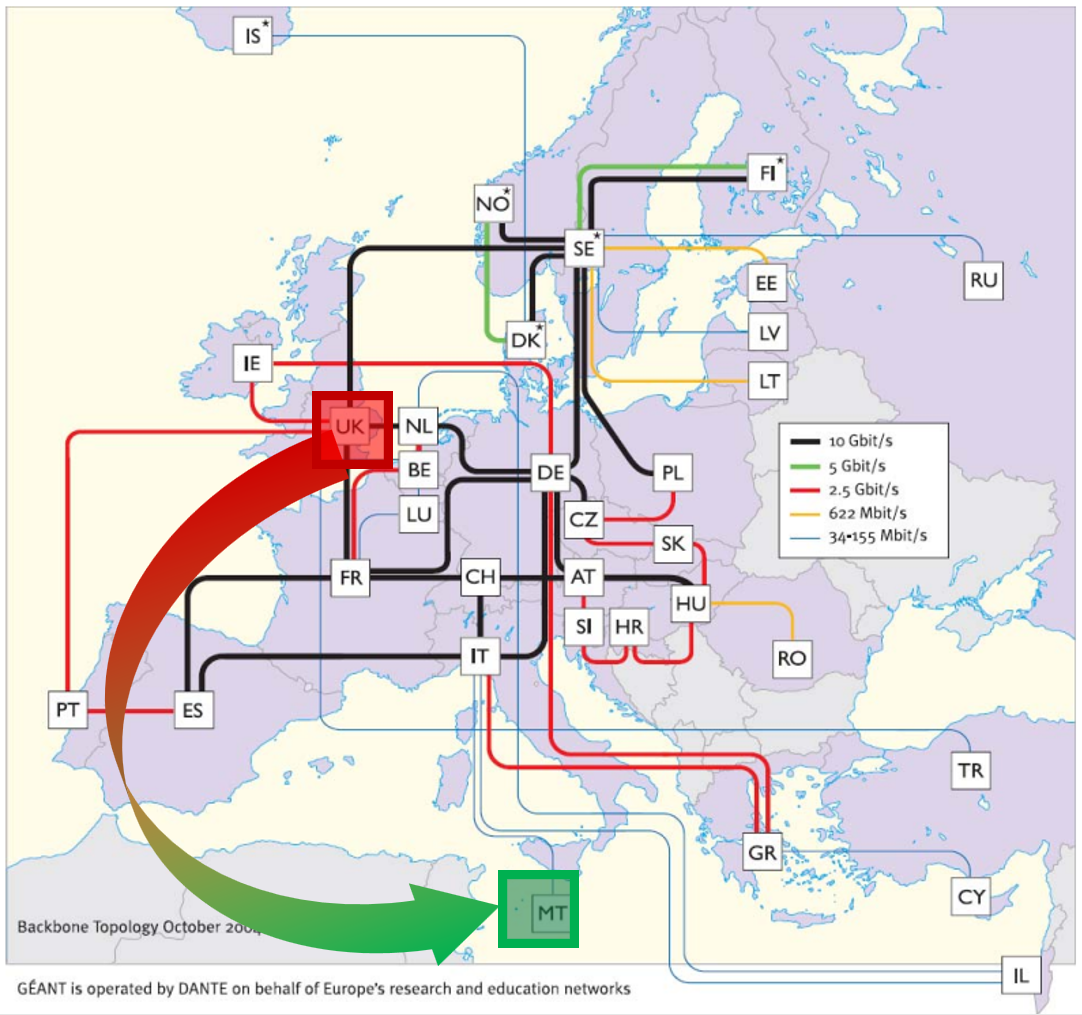}
            }%
    \centering
    \caption{Each picture represents a geographic adjacency map of the OD matrix in the corresponding dataset, each arrow represents an OD pair from the origin to the destination.}	\vspace{-0.5cm}
    \label{fig:1}
\end{figure*}

Origin-Destination (OD) matrices are widely used to provide location-based services (LBSs) or to aid in data mining and discovery in various scenarios related to geographic information. Specifically, OD matrices help to estimate travel expense \cite{li2018multi} and schedule  trips, so Google Maps provide OD matrices for better developing  LBSs. From a broader perspective, OD matrices support various data analysis tasks by introducing  geographic information. For instance, the COVID-19 spread in a country could be forecasted according to the  human movements OD matrix \cite{jiang2021countrywide}.
Moreover, internet traffic OD matrices show the potential to support  IP backbone network management \cite{liu2014prediction}, and internet resource allocation \cite{soule2005traffic}. In a word, the specific meaning of data stream is diverse between different application backgrounds, but we can still unify them into one cross-scenario application. Generally speaking, OD matrices record every traffic flow from origin regions to destination regions which are geographically divided. The regions of OD matrices could be city spans, country spans, or even entire continents spans. Each element $(i,j)$ in OD matrices is an OD pair that represents a certain traffic flow from region $i$ to region $j$ against a diverse background, as illustrated in Figure \ref{fig:1}. 

Here, we consider  the Origin-Destination matrix forecasting problem (ODMF) in which the elements represent traffic data for different application backgrounds (e.g., transportation traffic, IP backbone network traffic). There are many application scenarios for ODMF results. For example, in the field of transportation, it can be used to plan the itinerary in advance, the allocation of traffic resources, and in the field of IP network, it can be used to help efficiently complete tasks such as network planning or operation and maintenance.

\begin{figure}[ht]
    \centering
   \includegraphics[width=0.4\textwidth]{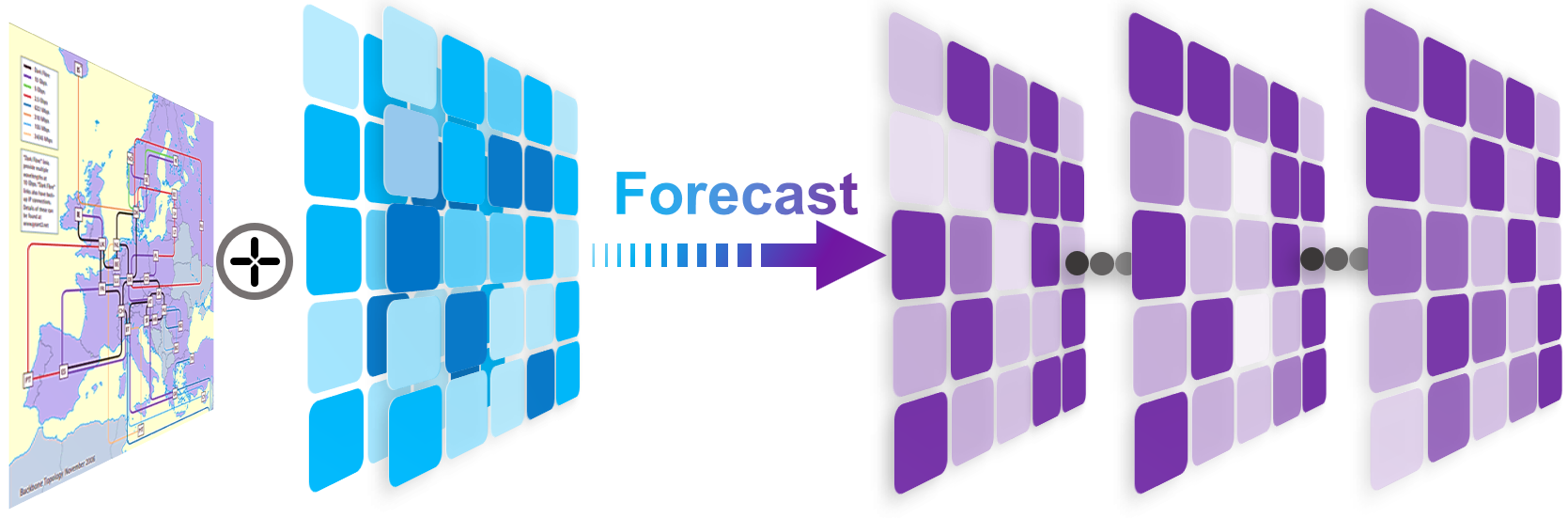}
    \caption{Long sequence OD matrices forecasting process  }	\vspace{-0.5cm}
    \label{fig:2}
\end{figure}

Existing methods are designed against  exceedingly fixed application backgrounds and under limited problem settings. On the one hand, those methods are specific to a single application \cite{hu2020stochastic,jiang2021countrywide,shi2020MPGCN},  thus, all the algorithm details intend to achieve the best performance on a single application scenario which leads to poor portability.
Further, they are invalid for revealing the intrinsic spatial-temporal features in OD matrix-type data across different application scenarios.
On the other hand, many methods \cite{jiangconv_LSTM,2017STGCN,zhao2020Germany50} design the experiment setting just to fulfill literally excellent results but ignore the actual demands of ODMF, such as long sequence time-series forecasting (LSTF). 
As the sampling interval is around 5 to 20 minutes in most OD matrix datasets, existing methods only predict the next several or even one timestep, which means predicting the next several minute's OD matrices. 
For resource allocation or another downstream task, this kind of prediction goes for nothing in a realistic application.
Therefore, we focus on the long sequence ODMF problem, which could be defined as follows: in multiple scenarios, based on OD matrices at several historical timesteps, we predict  OD matrices at longer future timesteps, illustrated in Figure \ref{fig:2}.

To address the above problems, we present a  spatial-temporal transformer (ODformer) for long sequence OD matrices forecasting. To the best of our knowledge, 
it is the first time to introduce the transformer to the ODMF problem, meanwhile with targeted improvements to accommodate the problem. 
We propose a  more general spatial dependency module that consists of the OD Attention mechanism and the 2D graph neural network, which captures spatial dependencies in multiple scenarios.
Further, we redesign the self-attention mechanism of the transformer named as PeriodSparse Self-attention to reduce the computational complexity of the long-sequence prediction tasks, while adapting to time series with diverse features from different scenarios. The contribution of this paper are summarized as follows:

\begin{itemize}
    \item ODformer is the first  Transformer-like model for the ODMF problem. Meanwhile, for the first time, it crosses multiple scenarios in ODMF, covering three real-world application scenarios: transportation traffic, crowd flow, and IP backbone network traffic.
    
    \item We propose OD Attention mechanism to mine the specialized spatial dependency between OD pairs of the same origin (destination), which is dramatically different from existing ODMF methods (i.e., merely considering the topological correlation between OD regions). Based on this, we achieve significant improvement in capturing spatial-temporal dependency in various application scenarios.
    
    \item To support long sequence forecasting, we propose PeriodSparse  Self-attention mechanism to capture long-range temporal correlation by achieving sparse attention with $\mathcal{O}(L\ln L)$ time complexity  and adapting to periodicity in different scenarios.

\end{itemize}
\section{Related works }
\subsection{OD Matrices Forecasting in Specific Applications }
We consider ODTF problems under different scenarios in turn: transportation traffic \cite{2017STGCN,li2018multi,dai2018short,GEML2019,wang2019simple,shi2020MPGCN,hu2020stochastic},
crowd flow \cite{jiang2019deepurbanevent,jiang2021countrywide}, IP backbone network traffic \cite{STTC2015,yang2020mstnn,zhao2019wavelet,jiangconv_LSTM,zhao2020Germany50}. Below, we will analyze the sophisticated differences  between diverse application backgrounds in the ODTF problem.

In the transportation traffic forecasting scenario, the data reveal the highest relationships with the geographic factors. Thus the point is how to capture spatial dependency. The direct idea is to transfer the 2D convolution to the spatial domain, and extract the spatial dependency in a way similar to processing pictures. For instance, STGCN \cite{2017STGCN} formulates the traffic spatial-temporal prediction task on graph structures.  MDL \cite{MDL2019ZHANG} designs two sophisticated models for predicting node flow and
 edge flow, respectively, based on fully convolutional networks to integrate transportation geographic factors. Despite convolutional neural networks(CNNs) could capture simple adjacent dependency between regions, the depth topological relationship between geographical regions is different from that in Euclidean space extracted by CNN. With the proposal of Graph Neural Network (GCN) \cite{gcn2016Defferrard}, some graph-based models \cite{2017STGCN,hu2020stochastic}, capture the topological information to forecast vehicle traffic flow between regions by employing GCNs that work in a non-euclidean space. In summary, the existing methods in the background of transportation traffic mainly focus on capturing spatial dependency by extracting the topological adjacent information.
 
 In the crowd flow forecasting scenario, the available data reveal different features of transportation background. The crowd data \cite{jiang2019deepurbanevent} is usually collected from smart devices  carried by humans that provide GPS information. In this case, the crowd data is more fine-grained and shows less relevance to road traffic structure. Due to the more complex spatial dependency, merely depending on geographic adjacency is far from enough, ODCRN \cite{jiang2021countrywide} proposes the Dynamic Graph Constructor to capture highly temporal-dependent relationships between OD regions.
 
 In the IP backbone network traffic forecasting scenario, the spatial dependency varies from the aforementioned. As the topology structures of IP networks consist of artificially set network nodes, the spatial dependency is comparatively irrelevant  to  natural geographic factors. Most existing methods \cite{STTC2015,zhao2019wavelet,jiangconv_LSTM} treat the network traffic volumes between IP nodes as grid-like data, thus applying convolution neural networks to extract spatial patterns. MSTNN \cite{yang2020mstnn} simply transfers the graph-based method to the IP network data to try to capture the spatial dependency, just like in geography determined scenario, which ignores the particularity of artificial IP network nodes.
\subsection{Long Sequence Time-series Forecasting}
In the time-series forecasting problem, the ability to capture precise long-term dependency \cite{DSTP-RNN2020,Autoformer2021} between historical sequence and predicted sequence determines the availability degree against the real-world application. Long sequence time-series forecasting (LSTF) advocates a longer predicted sequence than historical \cite{Encoder-decode2014}.

The traditional sequence-to-sequence LSTF models mostly apply recurrent neural networks (RNNs) \cite{RNN1994} such as LSTM \cite{Sequencetosequence2014} or GRU \cite{GRU2014empirical} to capture temporal patterns. The multilayered LSTM \cite{Sequencetosequence2014} maps the variable length input sequence to a vector of fixed dimensionality, then decodes the relatively long target sequence from the vector. And the RNN Encoder–Decoder \cite{Encoder-decode2014} proposes a gated recursive convolutional neural network to avoid the model degrading extremely rapidly when dealing with relatively long sequences. But the above RNN methods perform ineffectiveness when the prediction sequence becomes longer, which is probably due to the intrinsic limitations of recurrent neural network architectures \cite{dual-stageAttentionRNN2017}. 

Inducement of attention mechanism to time series forecasting problem brings a chance to  process long sequence. Dual-Stage Attention-Based Recurrent Neural Network \cite{dual-stageAttentionRNN2017,DSTP-RNN2020} adaptively extracts relevant input features at each time step by referring to the previous encoder hidden state, then selects relevant encoder hidden states across all time steps.
With the amelioration of canonical Transformer architecture \cite{transformer2017} against locality-agnostics and memory bottleneck \cite{EnhancingTransformer2019}, addressing the  LSTF problem by transformers became possible.
Autoformer \cite{Autoformer2021} adopts the sparse versions of point-wise self-attention for long series efficiency, meanwhile settling down the information utilization bottleneck. Informer \cite{informer2021}  designed a  ProbSparse Self-attention mechanism, which achieves $\mathcal{O}(L log L)$ in time complexity and memory usage.

\section{Problem Definition}
At first, we define several data forms that are crucial in the long sequence ODMF problem while trying our best to preserve generality.

As we consider the ODMF problem against diverse application scenarios, the raw data  varies in forms which  mostly consist of two types: traffic trajectories  and traffic matrices. 
\paragraph{Definition 1}(Origin-Destination Pair and Origin-Destination Matrix):
Typically, a trajectory(e.g. taxi trajectory, bus trajectory, human trajectory) is a time sequence while each timeslot contains a longitude-latitude coordinate that is denoted as: \\$[(t_{1}, l_{1}),(t_{2}, l_{2}), \ldots,(t_{n}, l_{n})]$, where each location $l$ is represented by a longitude-latitude coordinate and $t$ is the timestep. To obtain OD pairs, we segment each trajectory as follows: 
\begin{equation}
[(t_{1}, l_{1}),(t_{2}, l_{2}), \ldots,(t_{n}, l_{n})] \xrightarrow{\text { segment }}[\left\{(o_{1}, d_{1}),t_{1})\right\}, \ldots,\left\{(o_{n'}, d_{n'}),t_{n'}\right\}],
\label{eq:1}
\end{equation}
where the trajectory is segmented to $n'$ OP pairs. o,d represent the origin and destination regions. For improving generality, we take the departure timeslot of each trajectory segment as the $t_{n'}$ of the  OD pairs,
the approximation mainly due to (i) mostly, the trajectory segment time interval is less than the interval between  OD matrices' timeslot. (ii) some trajectory data are the trip forms \cite{hu2020stochastic}, which means the trajectory contains only one departure time and one arrival time.

In some cases, the raw data is in the form of a traffic matrix (i.e. a set of OD pairs \cite{jiangconv_LSTM,shi2020MPGCN}), which is even easier.
After the division of the area into regions (i.e., districts in cities, provinces in countries, IP internet  nodes), we could assign each OD pair to the elements in the OD matrix at timeslot $\mathcal{T}$:
\begin{equation}
\mathbf{M}_{\tau}^{i j}=\left\{(o,d,t)  \mid o_{(location)} \in \mathbb{OR}_{i} \wedge d_{(location)} \in \mathbb{DR}_{j} \wedge t =\mathcal{T} \right\}
\label{eq:2}
\end{equation}
The OD matrix is defined as $\mathbf{M}_{(\mathcal{T})} \in \mathbb{R}^{N \times N^{\prime} \times F}$, where the first two dimensions range over the origin and destination regions (i.e. $\mathbb{OR}$, $\mathbb{DR}$). The last dimension is denoted as the feature of each OP pair that in most cases is a single scalar, but for the sake of particular cases (e.g. multiple features\cite{hu2020stochastic}), we reserve that dimension.
\paragraph{Definition 2}(long sequence ODMF):
After the above process, we have  the input (e.i. the historical sequence at length $I$): $[\mathbf{M}_{(\mathcal{T}-I+1)},\ldots,\mathbf{M}_{(\mathcal{T}-1)},\mathbf{M}_{(\mathcal{T})},]$, the output is the predicted sequence at length $O$: $[\mathbf{M}_{(\mathcal{T}+1)},\mathbf{M}_{(\mathcal{T}+2)},\ldots,\mathbf{M}_{(\mathcal{T}+O)},]$
\begin{equation}
[\mathbf{M}_{(\mathcal{T}-I+1)},\ldots,\mathbf{M}_{(\mathcal{T}-1)},\mathbf{M}_{(\mathcal{T})}] \xrightarrow{}[\mathbf{M}_{(\mathcal{T}+1)},\mathbf{M}_{(\mathcal{T}+2)},\ldots,\mathbf{M}_{(\mathcal{T}+O)}],
\label{eq:3}
\end{equation}
The definitions  of "long sequence" differ in different long sequence time-series forecasting  methods. The "long sequence" in this paper has two meanings:  (i) $O$ is significantly longer than $I$ as primary LSTF methods \cite{Sequencetosequence2014,informer2021,Autoformer2021} encourage. (ii) $I$ is relatively  longer than most settings in other methods \cite{jiang2021countrywide,2017STGCN,jiangconv_LSTM}.

\begin{figure*}[tb]
    \centering
   \includegraphics[width=1\textwidth]{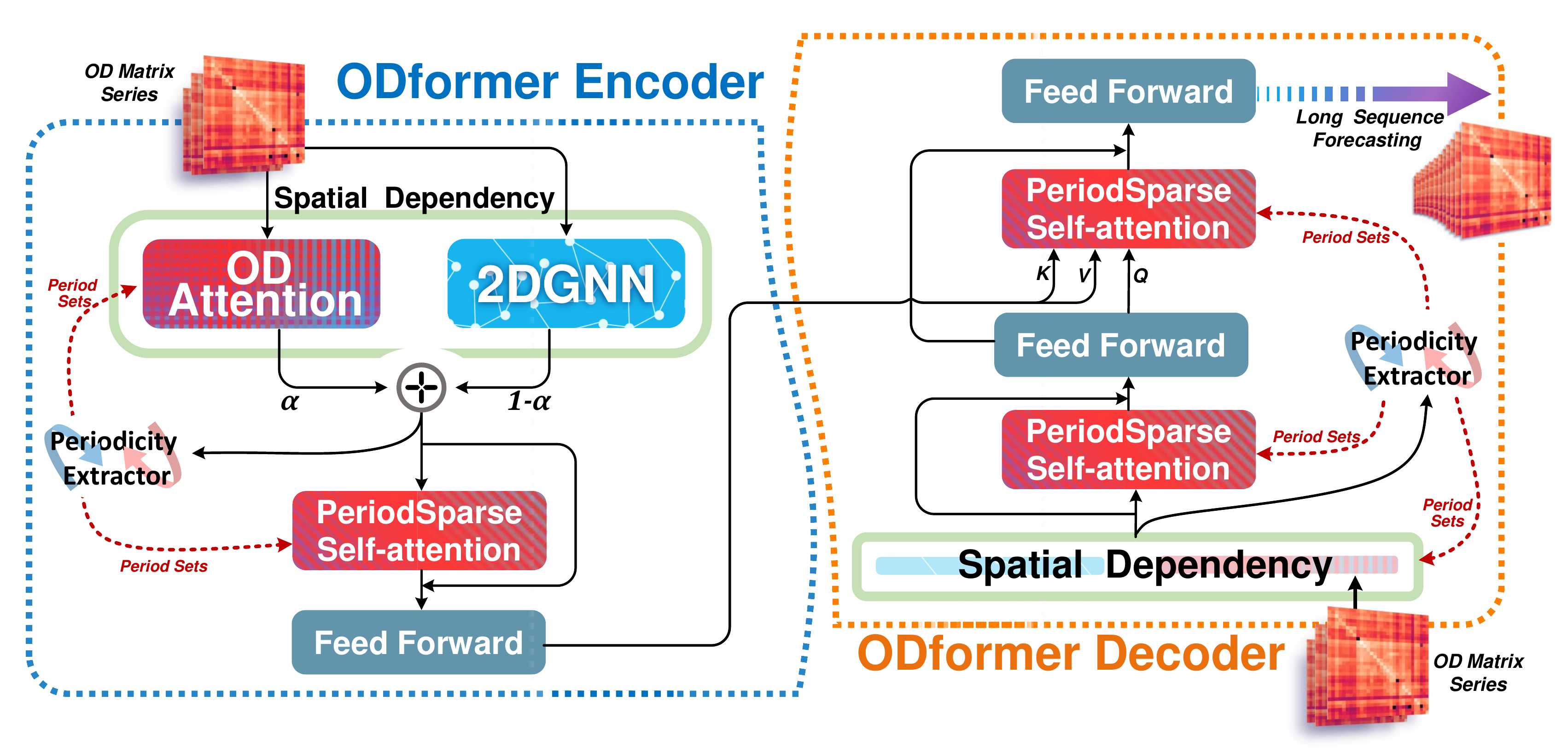}
    \caption{Architecture of ODformer }\vspace{-0.5cm}
    \label{fig:3}
\end{figure*}

\section{Methodology}
\subsection{Overall Architecture}
As shown in Figure \ref{fig:3}, we capture the spatial dependency before applying the attention mechanism in the temporal dimension. 
In each spatial dependency module, we apply the 2D-GNN and OD Attention mechanism
separately to capture spatial dependency upon 
different aspect (i.e., the adjacent relationships between regions and intrinsic dependency among the OD pairs which shares the same origin or destination). The output of the spatial dependency module is the aggregated results of the above two by the spatial share coefficient  $\alpha$, which varies according to different datasets. To complete sparse sampling for the two sparse attention mechanisms (i.e.,  OD Attention and PeriodSparse Self-attention), Periodicity Extractor dynamically compute the period sets with $k$ periods based on the output of the spatial dependency module. In this way, the output of the encoder contains both the spatial and temporal dependency of the OD matrix sequence, and finally, we complete the cross 
attention calculation together with the decoder.

\subsection{Spatial Dependency}
The spatial dependencies in different scenarios are quite different. For example, in transportation traffic, the spatial dependency is strongly correlated with geographic information, while in the IP network, the spatial dependencies are more manifested in the relationship between OD pairs. Based on this, we propose a spatial dependency module that combines two spatial dependencies mechanisms (i.e, OD Attention for dependencies between OD pairs and 2D-GCN for geographic dependencies between OD regions) to meet the needs of the model across scenarios: $\overline{M}_{\mathcal T}= (1-\alpha)M^\mathbb{G}_{\mathcal T}+\alpha M^\mathbb{A}_{\mathcal T}$, where the $M^\mathbb{G}_{\mathcal T}$, and $ M^\mathbb{A}_{\mathcal T}$ are the output of 2D-GCN and OD Attention. The shares of the two spatial dependencies are dynamically adjusted by the spatial share coefficient $\alpha$ according to different application scenarios.

\subsubsection{Origin-Destination (OD) Attention Mechanism: }Previous works \cite{hu2020stochastic,jiang2021countrywide,shi2020MPGCN,yang2020mstnn} capture the spatial correlation between origin regions or destination regions, but they roughly take all the OD pairs from the same origin or destination as the same while ignoring the dependency between OD pairs from the same regions.
\begin{figure*}[tb]
	\centering
		\begin{subfigure}{.49\textwidth}
			\centering\vspace{2.2cm}
			\includegraphics[width=0.6\textwidth]{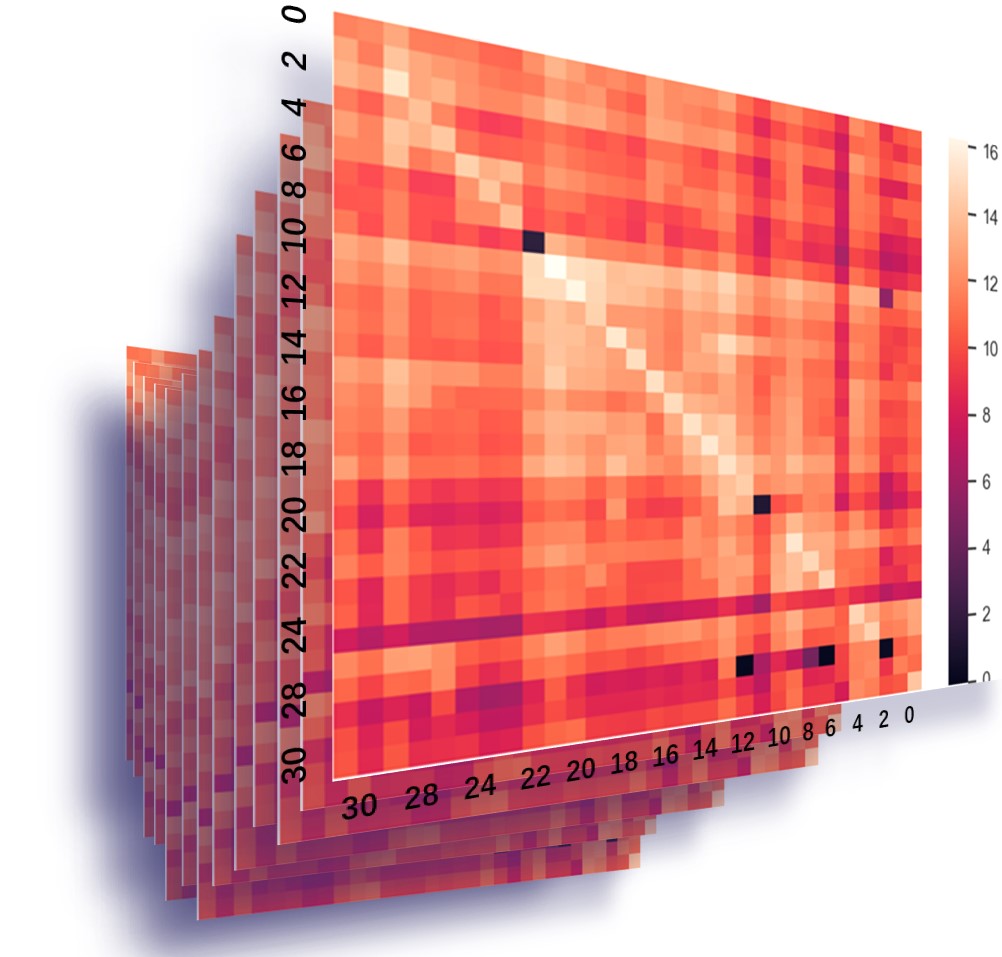}
			\caption{A OD matrix series(GÉANT)}
			\label{fig:4a}
		\end{subfigure}
		\begin{subfigure}{.49\textwidth}
			\centering
			\includegraphics[width=1\textwidth]{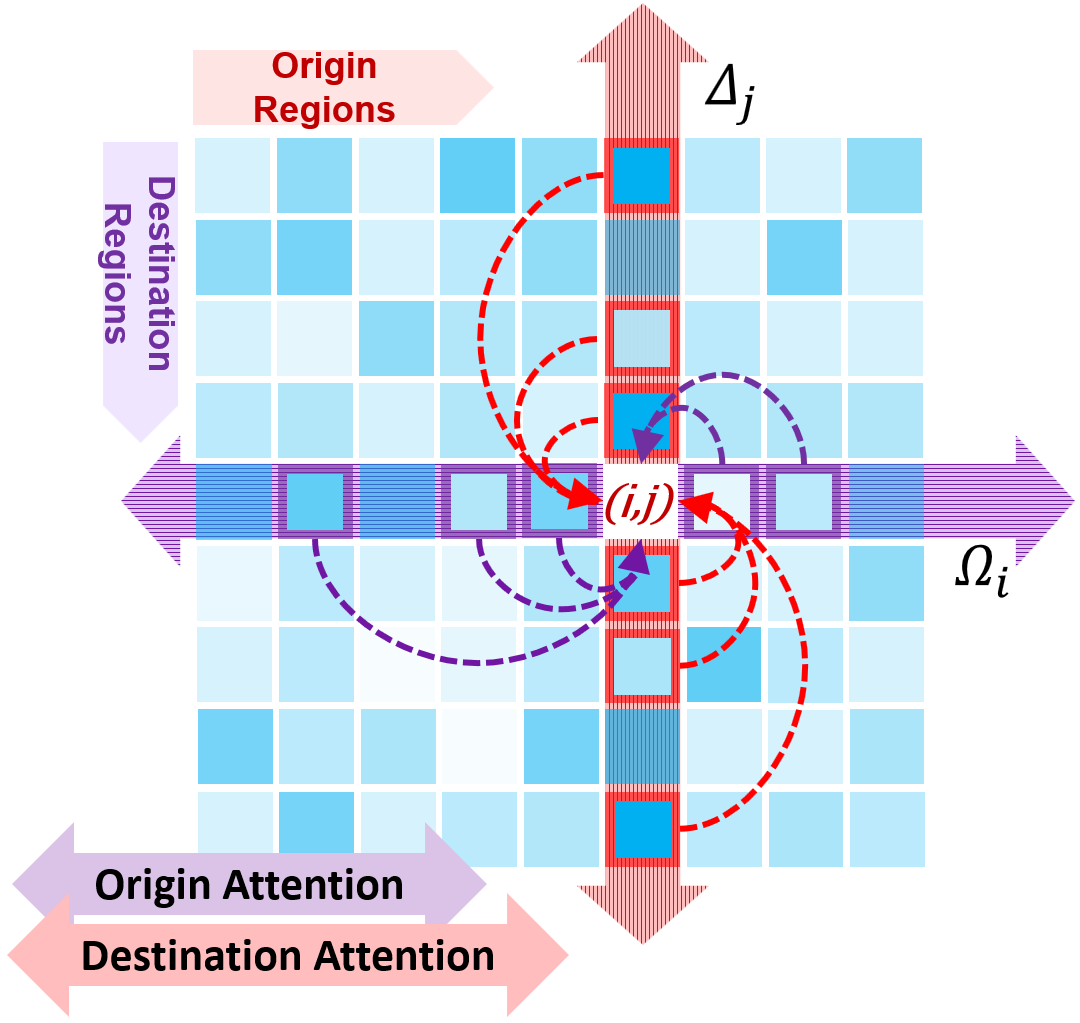}
			\caption{The OD Attention mechanism}
			\label{fig:4b}
		\end{subfigure}
  \caption{From the visualized OD matrices (a), we observe the bidirectional spatial dependence of each element (OD pair), which inspired us to propose OD Attention mechanism (b).}\vspace{-0.5cm}
\end{figure*}
Inspired by the spatial attention mechanism \cite{ASTGCN2019}, we propose the OD Attention mechanism to capture the dependencies between OD pairs of the same origin (destination). At timestep $\mathcal{T}$, the OD matrix: $\mathbf{M}_\mathcal{T} \in \mathbb{R}^{N \times N^{\prime} \times F}$ could be denoted as the origin  vector: $\mathcal{X}_\mathcal{T}=[\mathbf{X}_1, \mathbf{X}_2, \ldots, \mathbf{X}_N]^T, \mathbf{X}_n\in \mathbb{R}^{N^{\prime} \times F}$or destination  vector $\mathcal{Y}_\mathcal{T}=[\mathbf{Y}_1, \mathbf{Y}_2, \ldots, \mathbf{Y}_{N^{\prime}}] , \mathbf{Y}_n\in \mathbb{R}^{N \times F}$. To avoid the quadratic time complexity  on the region’s number  of the canonical self-attention \cite{vaswani2017attention}, which significantly restrains the capability in LSTF, here we propose a spatial sparse self-attention mechanism.
\paragraph{\textbf{(1) Sparse  Self-attention:}}
The brief strategy is to select the dominant query vector that contributes the most information.
We apply  the Shannon Entropy \cite{Shannon1991divergence} to select the one with the smallest value (e.g. the Shannon Entropy of uniform distribution is the highest, and the more  unlike the query  vector  and uniform distribution, the better. Since  the uniform  distribution is  to assign the even importance to each element, which is meaningless). Take the origin attention as an example, the i-th query's attention to all the keys is defined  in a probability form: $p\left(\mathbf{k}_{j} \mid \mathbf{q}_{i}\right)=e^{k\left(\mathbf{q}_{i}, \mathbf{k}_{j}\right)}/\sum_{j=1}^N {e^{ k\left(\mathbf{q}_{i}, \mathbf{k}_{
j}\right)}}$, where $k\left(\mathbf{q}_{i}, \mathbf{k}_{
j}\right)$ is the exponential kernel: $\exp \left(\frac{\mathbf{q}_{i} \mathbf{k}_{j}^{\top}}{\sqrt{d}}\right)$, thus the importance score  on each query $i$ is  defined below:
\begin{equation}
\mathbf I(\mathbf{q}_{i})=-\sum^N_{j=1}\left({p(\mathbf{k}_{j} \mid \mathbf{q}_{i})\sum^N_{j'\neq j}{k(\mathbf{q}_{i},\mathbf{k}_{j'})}}\right)
\end{equation}
According to the $\mathbf I(\mathbf{q}_{i})$, we choose the top $u$ smallest query vectors as the dominant vector to attend the attention computation and fill other elements with zeros.
To limit time consumption, we update the dominant query vector every $ P_{max}$ (the largest $ P_{k}$ from the period sets that will be illustrated next section) timesteps. Meanwhile, set $u = \ln N$, thus the total time complexity of OD Attention  is $\mathcal{O}(N\ln N)$. The sparse query vectors from $\mathcal {X _{T}, Y _{T}}$ is denoted as : $\mathcal {\overline{X_\mathcal{T}\mathbf{W}}}$, $\mathcal{ \overline{Y_{T}\mathbf{W}} }$.
\paragraph{\textbf{(2) Origin Attention:}} We perform the self-attention within the origin vectors to capture the dependencies between OD pairs derived from the same destination regions, the  origin  attention coefficients computes at below:
\begin{equation}
\mathbf{\varOmega^{\prime}_\mathcal{T}}= \sigma\left(\left(\overline{{\mathcal { {X} }}_\mathcal{T} \mathbf{W}_{1}}\right)^{T} \mathbf{W}_{2}\left(\mathbf{W}_{3} {\mathcal { X }}_\mathcal{T}\right)+\mathbf{b}_\varOmega\right)
\label{eq:4}
\end{equation}

\begin{equation}
\varOmega_{i, i'}=\frac{\exp \left(\varOmega^{\prime}_{i, i'}\right)}{\sum_{i'=1}^{N} \exp \left(\varOmega^{\prime}_{i, i'}\right)}
\label{eq:5}
\end{equation}
where $\mathbf{W}_{1}, \mathbf{W}_{2}, \mathbf{W}_{3}\in\mathbb{R}^{N\times N \times N'}$  are learnable parameter matrices, $\sigma$ is the activation function.
Equation \ref{eq:5} is meant to normalize the origin attention coefficient within each region.
The element $\varOmega_{i, i'}$ in the attention matrix reveals the relevance level between origin region $i$ to  another origin region $i'$.
\paragraph{\textbf{(3) Destination Attention:}}
In the almost same process, we symmetric compute the destination attention coefficient on top of $\mathcal{
\overline{ Y}_{T}}$ and $\mathcal{
{ Y}_{T}}$:
\begin{equation}
\mathbf{\varDelta^{\prime}_\mathcal{T}}= \sigma\left(\left(\overline{{\mathcal { Y }}_\mathcal{T} \mathbf{W}_{1}}\right) \mathbf{W}_{2}\left(\mathbf{W}_{3} {\mathcal { Y }}_\mathcal{T}\right)^{T}+\mathbf{b}_\varDelta\right)
\label{eq:6}
\end{equation}

\begin{equation}
\varDelta_{j, j'}=\frac{\exp \left(\varDelta^{\prime}_{j, j'}\right)}{\sum_{j'=1}^{N'} \exp \left(\varDelta^{\prime}_{j, j'}\right)}
\label{eq:7}
\end{equation}
At last, we apply the origin attention and destination attention matrix  to the OD matrix: $\mathbf{ M^\mathbb{A}_{\mathcal T}}=\varOmega_\mathcal{T}\mathbf{M}_\mathcal{T}\varDelta_\mathcal{T}$. Therefore, each element in the OD matrices is dynamically given the impact of the elements from the same origin or destination regions as illustrated in Figure \ref{fig:4b}.

\begin{figure*}[ht]
	\centering
		\begin{subfigure}{.3\textwidth}
			\centering
			\includegraphics[width=0.5\textwidth]{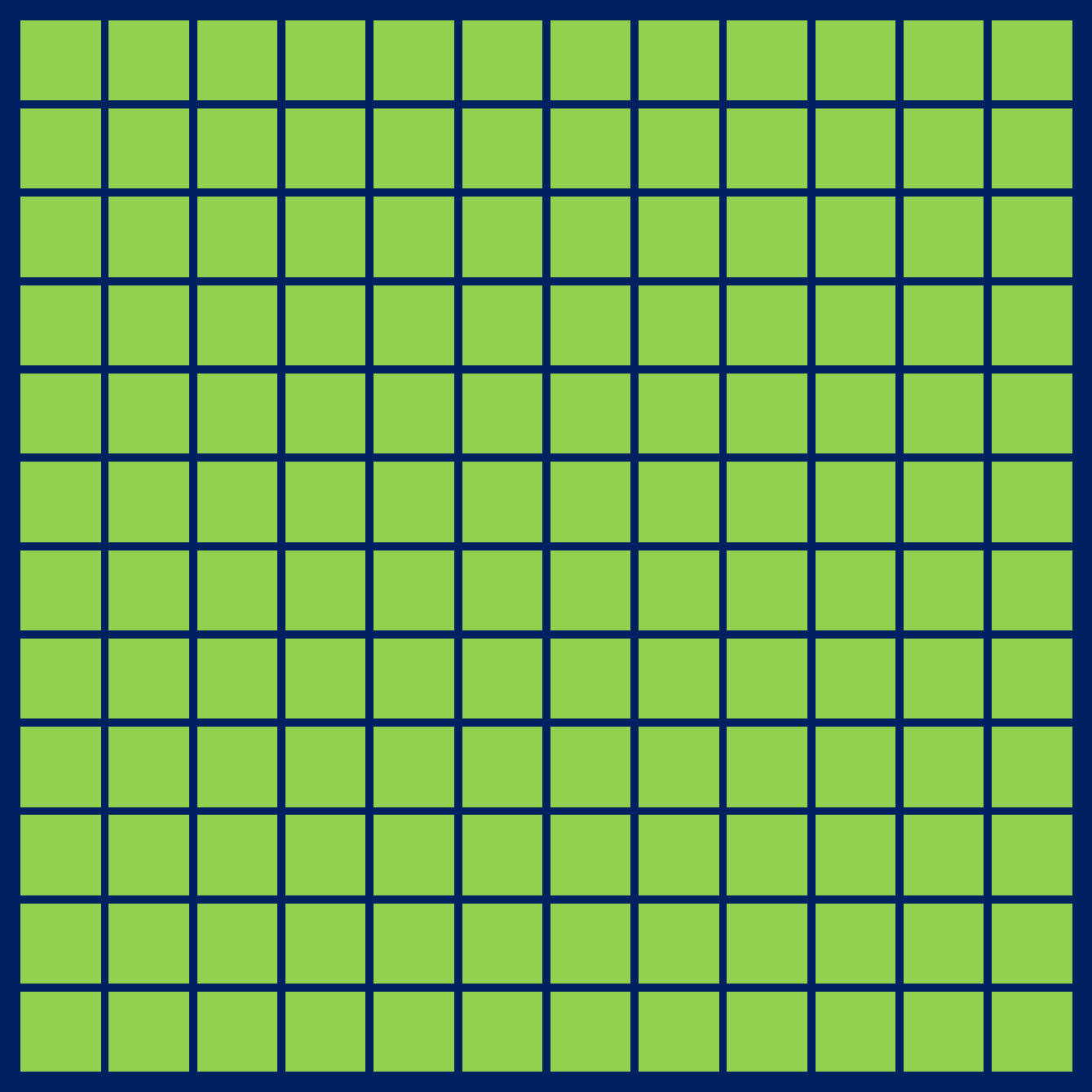}
			\caption{Full $\mathcal{O
			}(L^2)$ attention}
			\label{fig:5a}
		\end{subfigure}
		\begin{subfigure}{.3\textwidth}
			\centering
			\includegraphics[width=0.5\textwidth]{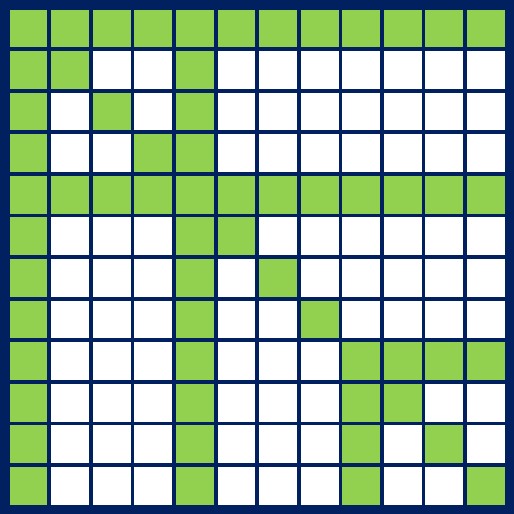}
			\caption{Fixed sparse}
			\label{fig:5b}
		\end{subfigure}
		\begin{subfigure}{.3\textwidth}
			\centering
			\includegraphics[width=0.5\textwidth]{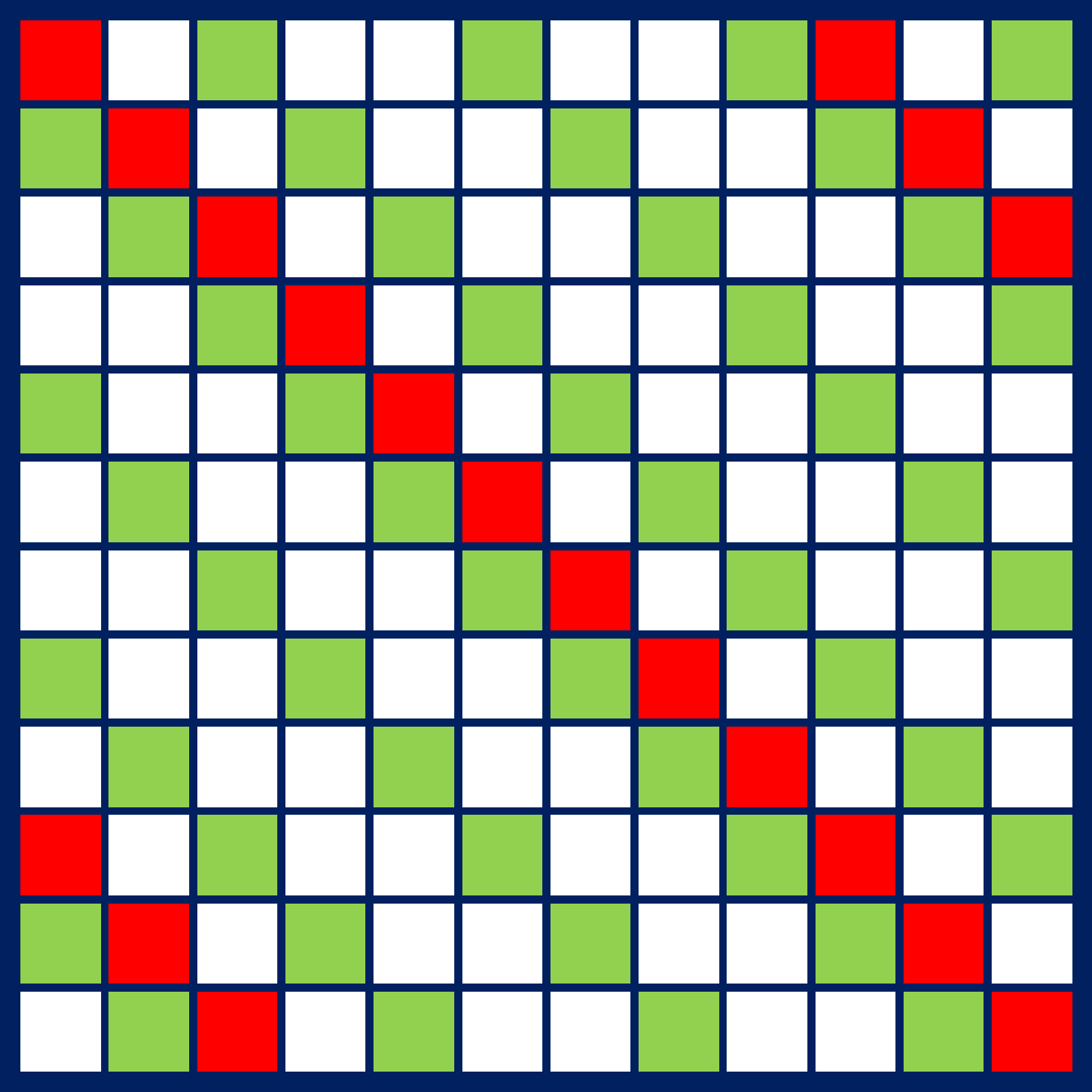}
			\caption{PeriodSparse }
			\label{fig:5c}
		\end{subfigure}
		
  \caption{The attention schemes shown in connectivity matrices.}\vspace{-0.5cm}
\end{figure*}

\subsubsection{2D Graph Convolution Network: }
Besides the intrinsic dependency between OD pairs, the topological relationship among diverse OD regions still plays a large or small role according to the different application scenarios.
In ODMF problems, graph convolution neural networks (GCNs) have wildly used to study the connectivity and globality based on topology graphs of regions. 

 Here, we adopt MGCNN \cite{GeometricMatrixCompletion2017} to capture the above relationships, which is an extended two-dimensional transformation of the GCN \cite{chebyNet2016} with Chebyshev polynomial approximation.
\begin{equation}
\mathbf{H}^{(l+1)}=\sigma\left(\sum_{i=0}^{K-1} \sum_{j=0}^{K-1}  T_{i}\left(\mathbf{L}_{origin}\right) \times \mathbf{H}^{(l)} \times T_{j}\left(\mathbf{L}_{destination}\right) \times \mathbf{W}_{\mathrm{ij}}\right)
\end{equation}
where  $\mathbf{L}_{origin}$ or $\mathbf{L}_{destination}$ is the Laplacian matrix obtained from $\mathbf{L}=\mathbf{I}-\mathbf{D}^{-1 / 2} \mathbf{A D}^{-1 / 2}$, $\mathbf{D}$ is the degree matrix, $\mathbf{I}$ is the identity matrix. $T_{i}\left(\mathbf{L}\right)$ is the Chebyshev polynomial of order $i$. $\mathbf{H}^{(l+1)}$ is the hidden state of layer $l$ while $\mathbf{H}^{(0)}$ is the OD matrix. After the 2D-GCN operation, we get the OD matrices: $\mathbf{ M^\mathbb{G}_{\mathcal T}}$ in Which the neighbor information propagates $
K$ hops between origin regions or destination regions.

\subsection{Temporal Dependency }
Long sequence forecasting of OD matrices mainly faces two challenges: (i) the intricate temporal pattern across the OD matrices. (ii) computation efficiency, which is even more matter due to the data scale of the OD matrices. Thus, we propose the Period Sparse Self-attention, which targets the above issue that captures intrinsic temporal dependency and reduces computation consumption at the same time. However, the periodicity in the differences scenario dataset differs a lot (e.g. the sampling interval is 15mins in the GÉANT dataset while 1 day in the Japan Human Trajectory dataset). For generality against different scenarios and to capture deep temporal dependency, we propose the Periodicity Extractor to adaptively give the period set: $\mathcal{P}=\{P_{1}, \cdots, P_{k}\}$ of the inputs sequence.

\subsubsection{Periodicity Extractor:}
For the purpose of adaptively extracting the period length of different datasets properly, inspired by STL \cite{STL1990}, we filter the tendency component. Given an OD  matrix sequence that has gone trough the spatial dependency module: \\$\mathcal{S }=[\mathbf{\overline {M}}_{(\mathcal{T}-s+1)},\ldots,\mathbf{ \overline{M}}_{(\mathcal{T}-1)},\mathbf{\overline{M}}_{(\mathcal{T})},]$, the periodicity component is computed as follows.
\begin{equation}
\begin{aligned}
&\mathcal{S}_{\mathrm{trend}}=\operatorname{MovAvr_{win}}(\mathcal{S}) \\
&\mathcal{S}_{\mathrm{period}}=\mathcal{S}-\mathcal{S}_{\mathrm{trend}}
\end{aligned}
\end{equation}
we apply centered  moving average (window size is $\operatorname{MovAvr_{win}}$ )to get the tendency
component, the remainder component is omitted for simplicity. For each sequence of length $L$:
$\mathcal{S_T}$($\mathcal T$  is the last timeslot ), we adopt the Auto-Correlation Mechanism \cite{Autoformer2021} to evaluate the chance of periodicity being $P$ based on it's periodicity component: $\mathcal{\hat{S}_T}$.
\begin{equation}
\begin{aligned}
&\mathcal{R}_{\mathcal{\hat{S}_T}}(P)=\lim _{L \rightarrow \infty} \frac{1}{L} \sum_{t=1}^{L} \mathcal{\hat S}_{\mathcal T} \mathcal{\hat S}_{\mathcal T-P}\\
&\mathcal{P}=\{P_{1}, \cdots, P_{k}\}=\underset{P \in\{1, \cdots, L\}}{ \operatorname{Topk}}\left(\mathcal{R}_{\mathcal{\hat{S}_T}}(P)\right)
\end{aligned}
\end{equation}
where  $\mathcal{R}_{\mathcal{\hat{S}_T}}(P)$ is the time-delay similarity between $\mathcal{\hat S_T}$ and its $P$ lag series $\mathcal{\hat S}_{\mathcal T-P}$ that could be calculated by Fast Fourier Transforms (FFT) \cite{wiener1930generalized}. $\operatorname{Topk}(\cdot)$ meant to get the periods of top $k$ time-delay similarity.

\subsubsection{PeriodSparse Self-attention:}Unlike the  canonical self-attention  mechanism (Figure \ref{fig:5a}) with $\mathcal{O}(L^2)$ time complexity or previous sparse attention that neglect  different data time interval (Figure \ref{fig:5b}), we propose the PeriodSparse Self-attention mechanism (Figure \ref{fig:5c})
to adaptively choose the sparse scheme according to different data.
Here  we define the connectivity pattern  $\mathcal{I}=\{I_1,\dots,I_n\} $, where  $I_i$ is the set of
the input vector indices which attends the $i$th output vector. Unlike the full self-attention  that allows every element to attend to all positions, the PeriodSparse Self-attention is determined by the subsets of $I_i$: $A_{i}^{(m)} \subset\{j: j \leq i\}$. Based on the extracted period sets $\mathcal{P}=\{P_{1}, \cdots, P_{k}\}$:
\begin{equation}
A_{i}^{(k)}=\{j: (i-j) \bmod P_k=0\}
\end{equation}
Notice that we set the number of attention heads according to $k$ (i.e., the indices of periods) so that we capture the spatial dependency at different periodicity by different attention heads. For every single head, we compute the output vector at each position:
\begin{equation}
a\left(\mathcal S_{i},{A}_{i} \right)=\operatorname{softmax}\left(\frac{\left(W_{q} \mathcal{S}_{i}\right) K_{A_{i}}^{T}}{\sqrt{d}}\right) V_{A_{i}}
\end{equation}
where $K_{A_{i}}=\left(W_{k} \mathcal{S}_{j}\right)_{j \in A_{i}}$, $V_{A_{i}}=\left(W_{v} \mathcal{S}_{j}\right)_{j \in A_{i}}$ and $W_{q},W_{k},W_{v}$ are to transform an input $\mathcal{S}$ to query, key, or value while $d$ is
the inner dimension of the queries and keys. We set $k=c\cdot \operatorname{max}\{(k' \mid \sum_{i}^{K'}\frac{L}{P_{i}}\}\leq \ln L)\}$, which makes the layer memory usage of  PeriodSparse Self-attention maintains $\mathcal{O}(L\ln L)$, $c$ is the sparsity control factor.


\begin{table*}[tb]
\caption{Long sequence ODMF results on five datasets (three scenarios)}
\centering\small\setlength\tabcolsep{2pt}\renewcommand{\arraystretch}{1.3}
\begin{tabular}{c|c|ccc|ccc|ccc|ccc|ccc}
\midrule[1pt]
                             &                          & \multicolumn{3}{c|}{GÉANT(48)}                                                                                        & \multicolumn{3}{c|}{Abilene(96)}                                                                                      & \multicolumn{3}{c|}{\begin{tabular}[c]{@{}c@{}}JHT(7)\end{tabular}}                             & \multicolumn{3}{c|}{NYC(12)}                                                                                          & \multicolumn{3}{c}{CD(12)}                                                                                            \\ \cline{3-17} 
\multirow{-2}{*}{Methods}    & \multirow{-2}{*}{Metric} & 96                                    & 192                                   & 288                                   & 192                                   & 288                                   & 384                                   & 14                                    & 28                                    & 54                                    & 24                                    & 36                                    & 48                                    & 24                                    & 36                                    & 48                                    \\ \midrule[0.7pt]
                             & MSE                      & 2.653                                 & 2.928                                 & 3.527                                 & 1.321                                 & 1.431                                 & 1.593                                 & 0.239                                 & 0.251                                 & 0.417                                 & 1.046                                 & 1.483                                 & 1.639                                 & 0.923                                 & 1.358                                 & 1.621                                 \\
\multirow{-2}{*}{LSTM\cite{LSTM1997}}       & MAE                      & 1.538                                 & 1.742                                 & 1.993                                 & 0.825                                 & 0.923                                 & 1.062                                 & 0.335                                 & 0.413                                 & 0.603                                 & 0.792                                 & 0.834                                 & 0.986                                 & 0.552                                 & 0.782                                 & 0.994                                 \\ \hline
                             & MSE                      & 2.033                                 & 2.972                                 & 3.039                                 & 1.175                                & 1.285                                 & 1.330                                 & 0.197                                 & 0.237                                 & 0.309                                 & 0.909 & 1.192                                 & 1.346                                 & 0.892                                 & 0.944  &0.951                               \\
\multirow{-2}{*}{ARIMA\cite{arima2014ariyo}}      & MAE                      & 1.239                                & 1.487                                 & 1.623                                 & 0.623                                 & 0.678                                 & 0.685                                & 0.303                                 & 0.364                                 & 0.478                                 & 0.512 & 0.578 & 0.693                                 & 0.238                                 & 0.273                                 & 0.286                                 \\ \hline
                             & MSE                      & 2.677                                 & 2.980                                 & 3.203                                 & 1.205                                 & 1.219                                 & 1.279                                 & 0.204                                 & 0.238                                 & 0.321                                 & 0.927 & 0.958 & 1.402                                 & 0.864                                 & 0.931                                 & 0.947                                 \\
\multirow{-2}{*}{N-BEAT\cite{N-BEAT2019nSoreshki}}      & MAE                      &1.502 &1.536 &1.854                                 &0.637 &0.649 &0.672 	&0.311 &0.372 &0.486	&0.527 &0.568 &0.699 	&0.234 &0.259 &0.277                                \\ \hline
                             & MSE                     
&1.924 &2.072 &2.133 	&{\color[HTML]{6200C9} \textbf{0.955}} &1.062 &1.298 	&0.240 &0.251 &0.271	 &0.947 &1.090 &1.204 	&0.943 &0.962 &0.998                                \\
\multirow{-2}{*}{DeepAR\cite{DeepAR2020salinasdeepar}}      & MAE                      &0.962 &1.003 &1.138 	&{\color[HTML]{6200C9} \textbf{0.591}} &0.619 &0.637 	&0.339 &0.352 &0.388 	&0.602 &0.637 &0.697 	&0.267 &0.291 &0.303
                              \\ \hline
                             & MSE                      & 2.498                                 & 2.832                                 & 3.317                                 & 1.289                                 & 1.299                                 & 1.324                                 & 0.199                                 & 0.242                                 & 0.313                                 & {\color[HTML]{6200C9} \textbf{0.836}} & {\color[HTML]{6200C9} \textbf{0.921}} & 1.237                                 & 0.743                                 & 0.817                                 & 0.864                                 \\
\multirow{-2}{*}{STGCN\cite{2017STGCN}}      & MAE                      & 1.490                                 & 1.528                                 & 1.734                                 & 0.679                                 & 0.690                                 & 0.698                                 & 0.317                                 & 0.385                                 & 0.492                                 & {\color[HTML]{6200C9} \textbf{0.492}} & {\color[HTML]{6200C9} \textbf{0.537}} & 0.684                                 & 0.216                                 & 0.223                                 & 0.229                                 \\ \hline
                             & MSE                      & 2.302                                 & 2.538                                 & 2.791                                 & 1.148                                 & 1.203                                 & 1.239                                 & 0.196                                 & 0.239                                 & 0.298                                 & 0.912                                 & 1.194                                 & 1.303                                 & {\color[HTML]{CB0000} \textbf{0.702}} & {\color[HTML]{6200C9} \textbf{0.791}} & {\color[HTML]{6200C9} \textbf{0.822}} \\
\multirow{-2}{*}{GEML\cite{GEML2019}}       & MAE                      & 1.173                                 & 1.297                                 & 1.418                                 & 0.647                                 & 0.682                                 & 0.691                                 & 0.310                                 & 0.371                                 & 0.472                                 & 0.523                                 & 0.602                                 & 0.684                                 & {\color[HTML]{CB0000} \textbf{0.201}} & {\color[HTML]{6200C9} \textbf{0.213}} & {\color[HTML]{6200C9} \textbf{0.226}} \\ \hline
                             & MSE                      & 2.587                                 & 2.781                                 & 2.970                                 & 1.238                                 & 1.246                                 & 1.301                                 & {\color[HTML]{6200C9} \textbf{0.178}} & {\color[HTML]{6200C9} \textbf{0.211}} & {\color[HTML]{000000} 0.276}          & 0.962                                 & 1.228                                 & 1.318                                 & 0.836                                 & 0.903                                 & 0.967                                 \\
\multirow{-2}{*}{ODCRN\cite{jiang2021countrywide}}      & MAE                      & 1.395                                 & 1.452                                 & 1.672                                 & 0.630                                 & 0.652                                 & 0.662                                 & {\color[HTML]{6200C9} \textbf{0.307}} & {\color[HTML]{6200C9} \textbf{0.337}} & {\color[HTML]{000000} 0.404}         & 0.591                                 & 0.611                                 & 0.697                                 & 0.241                                 & 0.269                                 & 0.293                                 \\ \hline
                             & MSE                      & {\color[HTML]{6200C9} \textbf{1.741}} & {\color[HTML]{000000} 2.088}          & {\color[HTML]{000000} 2.365}          & 1.132                                 & 1.168                                 & 1.230                                 & {\color[HTML]{000000} 0.184}          & {\color[HTML]{000000} 0.216}          & {\color[HTML]{000000} 0.283}          & 0.923                                 & 1.273                                 & 1.582                                 & 0.791                                 & 0.826                                 & 0.839                                 \\
\multirow{-2}{*}{MPGCN\cite{shi2020MPGCN}}      & MAE                      & {\color[HTML]{6200C9} \textbf{0.933}} & {\color[HTML]{000000} 1.085}          & {\color[HTML]{000000} 1.259}          & 0.631                                 & 0.655                                 & 0.669                                 & {\color[HTML]{000000} 0.310}          & {\color[HTML]{000000} 0.347}          & {\color[HTML]{000000} 0.413}          & 0.597                                 & 0.623                                 & 0.709                                 & 0.218                                 & 0.226                                 & 0.238                                 \\ \hline
                             & MSE                      & 4.309                                 & 4.540                                 & 4.583                                 & {\color[HTML]{000000} 1.173}          & {\color[HTML]{000000} 1.202}          & {\color[HTML]{000000} 1.269}          & 0.275                                 & 0.291                                 & 0.318                                 & 1.326                                 & 1.342                                 & 1.462                                 & 0.831                                 & 0.857                                 & 0.892                                 \\
\multirow{-2}{*}{Informer\cite{informer2021}}   & MAE                      & 1.518                                 & 1.693                                 & 1.783                                 & {\color[HTML]{000000} 0.611}          & {\color[HTML]{000000} 0.618}          & {\color[HTML]{000000} 0.632}          & 0.350                                 & 0.396                                 & 0.419                                 & 0.682                                 & 0.697                                 & 0.719                                 & 0.239                                 & 0.249                                 & 0.257                                 \\ \hline
                             & MSE                      & 1.855                                 & {\color[HTML]{6200C9} \textbf{1.986}} & {\color[HTML]{6200C9} \textbf{2.013}} & 0.971 & {\color[HTML]{6200C9} \textbf{0.992}} & {\color[HTML]{6200C9} \textbf{1.039}} & 0.236                                 & 0.239                                 & {\color[HTML]{6200C9} \textbf{0.250}} & 0.937                                 & 0.993                                 & {\color[HTML]{6200C9} \textbf{1.109}} & 0.910                                 & 0.929                                 & 0.942                                 \\
\multirow{-2}{*}{Autoformer\cite{Autoformer2021}} & MAE                      & 0.945                                 & {\color[HTML]{6200C9} \textbf{0.989}} & {\color[HTML]{6200C9} \textbf{0.992}} &0.602 & {\color[HTML]{6200C9} \textbf{0.610}} & {\color[HTML]{6200C9} \textbf{0.627}} & 0.327                                 & 0.341                                 & {\color[HTML]{6200C9} \textbf{0.352}} & 0.582                                 & 0.613                                 & {\color[HTML]{6200C9} \textbf{0.676}} & 0.252                                 & 0.278                                 & 0.281                                 \\ \hline
                             & MSE                      & {\color[HTML]{CB0000} \textbf{1.623}} & {\color[HTML]{CB0000} \textbf{1.735}} & {\color[HTML]{CB0000} \textbf{1.843}} & {\color[HTML]{CB0000} \textbf{0.936}} & {\color[HTML]{CB0000} \textbf{0.961}} & {\color[HTML]{CB0000} \textbf{0.997}} & {\color[HTML]{CB0000} \textbf{0.163}} & {\color[HTML]{CB0000} \textbf{0.186}} & {\color[HTML]{CB0000} \textbf{0.224}} & {\color[HTML]{CB0000} \textbf{0.815}} & {\color[HTML]{CB0000} \textbf{0.901}} & {\color[HTML]{CB0000} \textbf{0.977}} & {\color[HTML]{6200C9} \textbf{0.704}} & {\color[HTML]{CB0000} \textbf{0.788}} & {\color[HTML]{CB0000} \textbf{0.804}} \\
\multirow{-2}{*}{\textbf{ODformer}}   & MAE                      & {\color[HTML]{CB0000} \textbf{0.908}} & {\color[HTML]{CB0000} \textbf{0.937}} & {\color[HTML]{CB0000} \textbf{0.940}} & {\color[HTML]{CB0000} \textbf{0.578}} & {\color[HTML]{CB0000} \textbf{0.583}} & {\color[HTML]{CB0000} \textbf{0.602}} & {\color[HTML]{CB0000} \textbf{0.277}} & {\color[HTML]{CB0000} \textbf{0.310}} & {\color[HTML]{CB0000} \textbf{0.339}} & {\color[HTML]{CB0000} \textbf{0.484}} & {\color[HTML]{CB0000} \textbf{0.512}} & {\color[HTML]{CB0000} \textbf{0.578}} & {\color[HTML]{6200C9} \textbf{0.209}} & {\color[HTML]{CB0000} \textbf{0.210}} & {\color[HTML]{CB0000} \textbf{0.217}} \\ \midrule[1pt]
\end{tabular}
\label{table:1}
\end{table*}
\section{Experiments}
\subsection{Data}
We perform experiments on five datasets in three application scenarios (i.e., IP backbone network traffic, crowd flow, transportation traffic) to evaluate the predictive ability of the proposed model across scenarios.
\\
\textbf{IP backbone network traffic:} In this scenario, we apply two datasets: the GÉANT\footnote{https://network.geant.org/} dataset and the Abilene\footnote{https://math.bu.edu/people/kolaczyk/datasets.html} dataset. The former is collected from the Abilene network, including 11 nodes (regions) and 48,097 time intervals that differ by five minutes, and the latter is collected from the pan-European research and education network GÉANT, including 23 routing nodes (regions) and 10,772  time intervals that differ by 15 minutes. The input of data includes OD matrix series and network topology graphs between nodes (regions).
\\
\textbf{Crowd flow:} We adopt the Japan Human Trajectory (JHT)\footnote{https://github.com/deepkashiwa20/ODCRN/tree/main/data} dataset to evaluate the models in the crowd flow scenario. The OD regions are the 47 prefectures of Japan, the related topological map is the adjacency map of Japanese provinces, and the time sampling interval is one day (2020/1/1/ $\sim$ 2021/2/28).
\\
\textbf{Transportation traffic: }To evaluate the models in traffic scenarios, we adopt two taxi trajectory datasets: New York City (NYC)\footnote{https://www1.nyc.gov/site/tlc/about/tlc-trip-record-data.page} dataset, which records taxi trips collected from 2013-11-01 to 2013-12-31 from Manhattan (67 subregions), the time interval is 15 minutes. Chengdu  (CD)\footnote{https://goo.gl/3VsEym} dataset: records from 1.4 billion  taxi trips collected from 2014-08-03 to 2014-08-30 in Chengdu (79 regions), China.
\subsection{Experimental Details}
\subsubsection{Baselines: }We implement two classes of baselines to compare and evaluate our proposed model on the long sequence ODMF task: \\\textit{From the perspective of the ODMF problem: } (1) STGCN \cite{2017STGCN}: ST-GCN is one of the earliest models that induce graph convolution network (GCN) to the spatiotemporal transportation prediction problems. (2) GEML \cite{GEML2019}: One of the state-of-the-art ODMF models in transportation background. (3) MPGCN \cite{shi2020MPGCN}: The first model that applied 2D-GCN with multiple graphs to the ODMF problem.
(4) ODCRN \cite{jiang2021countrywide}: The state-of-the-art ODMF model in crowd flow background, with Dynamic Graph Constructor (DGC) module.
\\\textit{From the perspective of the LSTF problem: } (5)  LSTM \cite{LSTM1997}: One of the most common recurrent neural  networks. (6) ARIMA \cite{arima2014ariyo}: Classic benchmark model that captures seasonality adequately. (7) N-BEAT \cite{N-BEAT2019nSoreshki}, (8) DeepAR  \cite{DeepAR2020salinasdeepar}: Standard time series methods. (9) Informer \cite{informer2021}: One of the most advanced long sequence prediction models based on ProbSparse Self-attention mechanism. (10) Autoformer \cite{Autoformer2021}: An advanced long sequence prediction model with Auto-Correlation mechanism.
\begin{table*}[ht]
\centering\small\setlength\tabcolsep{3pt}\renewcommand{\arraystretch}{1.2}
\caption\normalsize{Comparison of PeriodSparse Self-attention and full
 self-attention.}
 \begin{threeparttable}
\begin{tabular}{cc|cccccccccccc}
\midrule[1pt]
\multicolumn{2}{c|}{Input length I}                                                                         & \multicolumn{3}{c|}{48}                                                                                               & \multicolumn{3}{c|}{96}                                                                                               & \multicolumn{3}{c|}{192}                                                                                              & \multicolumn{3}{c}{384}                                                                                               \\ \hline
\multicolumn{2}{c|}{prediction length O}                                                                    & 96                                    & 192                                   & \multicolumn{1}{c|}{384}              & 192                                   & 384                                   & \multicolumn{1}{c|}{768}              & 384                                   & 768                                   & \multicolumn{1}{c|}{1536}             & 768                                   & 1536                                  & 3072                                  \\ \midrule[0.7pt]
\multicolumn{1}{c|}{}                                                                                 & MSE & 0.978                                 & 1.194                                 & 1.397                                 & 0.999                                 & 1.078                                 & -                                     & 1.152                                    & -                                     & -                                     & -                                     & -                                     & -                                     \\ \cline{2-14} 
\multicolumn{1}{c|}{\multirow{-2}{*}{\begin{tabular}[c]{@{}c@{}}Full\\ Attention\end{tabular}}}       & MAE & 0.539                                 & 0.611                                 & 0.793                                 & 0.623                                 & 0.650                                 & -                                     & 0.690                                    & -                                     & -                                     & -                                     & -                                     & -                                     \\ \hline
\multicolumn{1}{c|}{}                                                                                 & MSE & {\color[HTML]{CB0000} \textbf{0.947}} & {\color[HTML]{CB0000} \textbf{0.981}} & {\color[HTML]{CB0000} \textbf{1.184}} & {\color[HTML]{CB0000} \textbf{0.936}} & {\color[HTML]{CB0000} \textbf{0.997}} & {\color[HTML]{CB0000} \textbf{1.359}} & {\color[HTML]{CB0000} \textbf{0.983}} & {\color[HTML]{CB0000} \textbf{1.172}} & {\color[HTML]{CB0000} \textbf{1.320}} & {\color[HTML]{CB0000} \textbf{1.086}} & {\color[HTML]{CB0000} \textbf{1.238}} & {\color[HTML]{CB0000} \textbf{1.492}} \\ \cline{2-14} 
\multicolumn{1}{c|}{\multirow{-2}{*}{\begin{tabular}[c]{@{}c@{}}ProbSparse\\ Attention\end{tabular}}} & MAE & {\color[HTML]{CB0000} \textbf{0.413}} & {\color[HTML]{CB0000} \textbf{0.489}} & {\color[HTML]{CB0000} \textbf{0.560}} & {\color[HTML]{CB0000} \textbf{0.578}} & {\color[HTML]{CB0000} \textbf{0.602}} & {\color[HTML]{CB0000} \textbf{0.673}} & {\color[HTML]{CB0000} \textbf{0.632}} & {\color[HTML]{CB0000} \textbf{0.687}} & {\color[HTML]{CB0000} \textbf{0.779}} & {\color[HTML]{CB0000} \textbf{0.951}} & {\color[HTML]{CB0000} \textbf{1.083}} & {\color[HTML]{CB0000} \textbf{1.170}} \\ \midrule[1pt]
\end{tabular}
\begin{tablenotes}\renewcommand{\arraystretch}{0.5}
            \item[1] We replace the PeriodSparse Self-attention in ODformer with full self-attention.
            \item[2]  The “-” indicates the out-of-memory.
\end{tablenotes}
\end{threeparttable}
\label{table:2}
\end{table*}
\begin{table*}[ht]
\centering\setlength\tabcolsep{5pt}\renewcommand{\arraystretch}{1.2}
\caption{Ablation of OD Attention or 2D-GCN module in ODformer.}\label{table:3}
\begin{threeparttable}
\begin{tabular}{c|c|cccccccccc}
\midrule[1pt]
                                                                                  &                          & \multicolumn{2}{c|}{GÉANT(48)}                                                & \multicolumn{2}{c|}{Abilene(96)}                                              & \multicolumn{2}{c|}{\begin{tabular}[c]{@{}c@{}}JHT(7)\end{tabular}} & \multicolumn{2}{c|}{YNC(12)}                                                  & \multicolumn{2}{c}{CD(12)}                                                    \\ \cline{3-12} 
\multirow{-2}{*}{Method}                                                          & \multirow{-2}{*}{Metric} & 96                                    & 192                                   & 192                                   & 288                                   & 14                                          & 28                                          & 24                                    & 36                                    & 24                                    & 36                                    \\ \midrule[0.7pt]
                                                                                  & MSE                      & {\color[HTML]{CB0000} \textbf{1.623}} & {\color[HTML]{CB0000} \textbf{1.735}} & {\color[HTML]{CB0000} \textbf{0.936}} & {\color[HTML]{CB0000} \textbf{0.961}} & {\color[HTML]{CB0000} \textbf{0.163}}       & {\color[HTML]{CB0000} \textbf{0.186}}       & {\color[HTML]{CB0000} \textbf{0.815}} & {\color[HTML]{CB0000} \textbf{0.901}} & {\color[HTML]{CB0000} \textbf{0.704}} & {\color[HTML]{CB0000} \textbf{0.788}} \\ \cline{2-12} 
\multirow{-2}{*}{ODformer}                                                        & MAE                      & {\color[HTML]{CB0000} \textbf{0.908}} & {\color[HTML]{CB0000} \textbf{0.937}} & {\color[HTML]{CB0000} \textbf{0.578}} & {\color[HTML]{CB0000} \textbf{0.583}} & {\color[HTML]{CB0000} \textbf{0.277}}       & {\color[HTML]{CB0000} \textbf{0.310}}       & {\color[HTML]{CB0000} \textbf{0.484}} & {\color[HTML]{CB0000} \textbf{0.512}} & {\color[HTML]{CB0000} \textbf{0.209}} & {\color[HTML]{CB0000} \textbf{0.210}} \\ \hline
                                                                                  & MSE                      & 1.690                                 & 1.867                                 & 0.955                                 & 0.971                                 & 0.169                                       & 0.198                                       & {\color[HTML]{6200C9} \textbf{0.819}} & {\color[HTML]{6200C9} \textbf{0.909}} & {\color[HTML]{6200C9} \textbf{0.710}} & {\color[HTML]{6200C9} \textbf{0.793}} \\ \cline{2-12} 
\multirow{-2}{*}{\begin{tabular}[c]{@{}c@{}}ODformer$^{\dagger}$\end{tabular}} & MAE                      & 0.926                                 & 0.951                                 & 0.589                                 & 0.599                                 & 0.286                                       & 0.317                                       & {\color[HTML]{6200C9} \textbf{0.490}} & {\color[HTML]{6200C9} \textbf{0.520}} & {\color[HTML]{6200C9} \textbf{0.212}} & {\color[HTML]{6200C9} \textbf{0.217}} \\ \hline
                                                                                  & MSE                      & {\color[HTML]{6200C9} \textbf{1.630}} & {\color[HTML]{6200C9} \textbf{1.742}} & {\color[HTML]{6200C9} \textbf{0.942}} & {\color[HTML]{6200C9} \textbf{0.966}} & {\color[HTML]{6200C9} \textbf{0.167}}       & {\color[HTML]{6200C9} \textbf{0.190}}       & 0.834                                 & 0.913                                 & 0.711                                 & 0.799                                 \\ \cline{2-12} 
\multirow{-2}{*}{\begin{tabular}[c]{@{}c@{}}ODformer$^{\S}$\end{tabular}}    & MAE                      & {\color[HTML]{6200C9} \textbf{0.912}} & {\color[HTML]{6200C9} \textbf{0.941}} & {\color[HTML]{6200C9} \textbf{0.585}} & {\color[HTML]{6200C9} \textbf{0.590}} & {\color[HTML]{6200C9} \textbf{0.282}}       & {\color[HTML]{6200C9} \textbf{0.311}}       & 0.497                                 & 0.527                                 & 0.214                                 & 0.223                                 \\ \midrule[1pt]

\end{tabular}
\begin{tablenotes}\renewcommand{\arraystretch}{0.5}
            \item[1] ODformer$^{\dagger}$ removes OD Attention from ODformer.
            \item[2] ODformer$^{\S}$ removes 2D-GCN from ODformer.
\end{tablenotes}
\end{threeparttable}
\end{table*}
\subsubsection{Setting: } We use a threshold of 98\% to clip overly large data to remove outliers. Then, fill in the missing values based on topological neighbors of the OD regions: \\$\mathbb{M}_{u,v} \leftarrow \operatorname{AGGREGATE} \left(\left\{\mathbf{M}_{\overline{u},v}\mid \forall \overline{u} \in \mathcal{N}(u)\right\}\cup \left\{\mathbf{M}_{u,\overline{v}}\mid \forall \overline{v} \in \mathcal{N}(v) \right\}\right)$, where $\operatorname{AGGREGATE}$ is the aggregator function \cite{GraphSAGE} that aggregates information from one-hop neighbors, $\mathcal{N}(v)$ denoted the immediate neighborhood of vertex $v$. To normalize the data, we take natural logarithm of the  elements in the original OD matrices. The training/validation/test dataset split ratio is: 6:2:2. The initial learning rate is set to  $10^{ - 4}$, using the ADAM \cite{adam2014} optimizer. The batch size is set to 16. The training process is early stopped within 8 epochs. All experiments are repeated five times, implemented in PyTorch \cite{Pytorch2019}, and conducted on a single NVIDIA GeForce RTX 3090 GPU. At last, we evaluate the overall performance of models on the long sequence ODMF by two metrics: MSE (Mean Square Error),  MAE (Mean Absolute Error).

\subsection{Results and Analysis}

\subsubsection{Overall performance: }Table \ref{table:1} compares the overall performance between the baselines and proposed ODformer on the long sequence ODMF task in data of three scenarios (The input length is on the right side of the dataset).
We gradually prolong the prediction length $O$ to verify the long sequence prediction ability of the model. The length setting depends on the specific scale of OD matrices so that long sequence ODMF is tractable on one single GPU for every method. Optimal and suboptimal results are shown in red and purple, respectively. For Table \ref{table:1}, we summarize as follows : (1) In the setting with the smallest prediction length, ODformer greatly improves the inference performance for vast majority scenarios, which illustrates the success of capturing spatial dependencies across scenarios. (2) The prediction error of ODformer rises smoothly and slowly over the growing prediction length $O$, further outperforming the rest of the models as the $O$ grows. This demonstrates ODformer's success in improving the predictive power towards long sequence OD matrices. (3)  The sub-optimal models in different scenarios are different, but our models can all perform optimally, which shows that ODformer successfully solves the problem of the ODMF cross scenario. (4) On the first set of Chengdu data, the performance of ODformer is slightly weaker than that of GEML, which may duo to the spatial relationships in this dataset all depending on the adjacency information between regions.

\begin{figure}[h]
	\centering
		\begin{subfigure}{0.25\textwidth}\vspace{-0.15cm}
			\centering
			\includegraphics[width=1\textwidth]{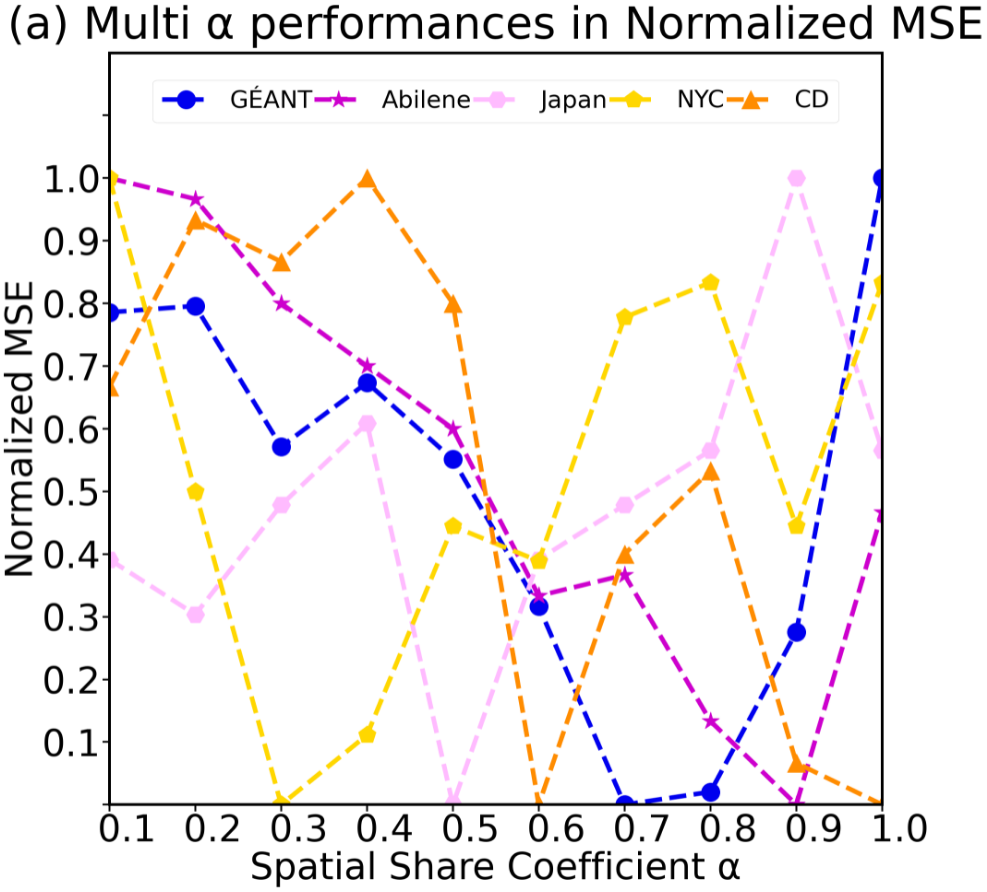}
			\label{fig:6a}
		\end{subfigure}
		\begin{subfigure}{.27\textwidth}
			\centering
			\includegraphics[width=1\textwidth]{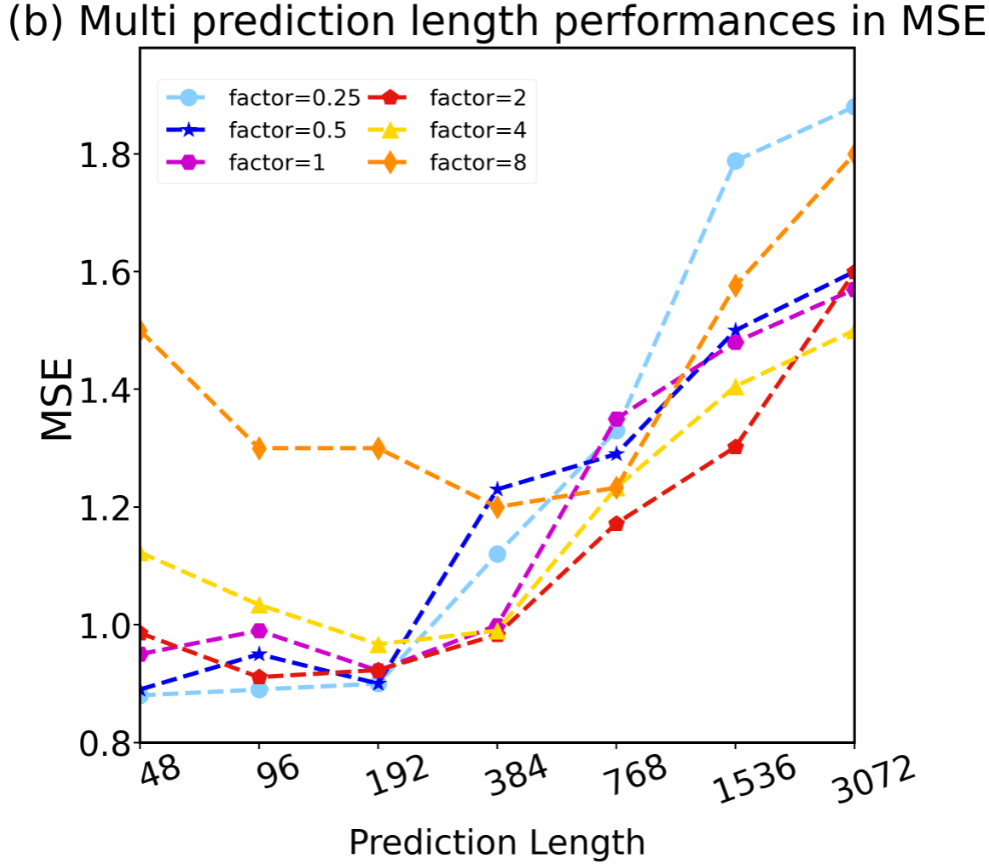}
			\label{fig:6b}
		\end{subfigure}
		\vspace{-0.5cm}
  \caption{The parameters sensitivity of ODformer.}	\vspace{-.5cm}
  \label{fig:6}
\end{figure}

\subsubsection{Ablation study:  }To demonstrate the effectiveness of our innovative modules, we performed ablation experiments as follows.\\
\textit{(1) Removal of PeriodSparse Self-attention: }The experiment was performed on the Abilene dataset. In Table \ref{table:2}, PeriodSparse Self-attention outperforms the full attention mechanism under various $O-I$ settings. At the same time, it also significantly outperforms full attention in memory consumption.\\
\textit{(2) Removal of OD Attention mechanism: }As shown in Table \ref{table:3}, ODformer with the full spatial dependency module performs the best on all datasets. However, we can find that the contribution of OD Attention is more dominant in IP backbone network traffic and crowd flow scenarios, while in the scenarios with more relevant geographic information, 2D-GCN contributes more.
\begin{figure}[h]
	\centering
		\begin{subfigure}{0.4\textwidth}\vspace{-0.15cm}
			\centering
			\includegraphics[width=1\textwidth]{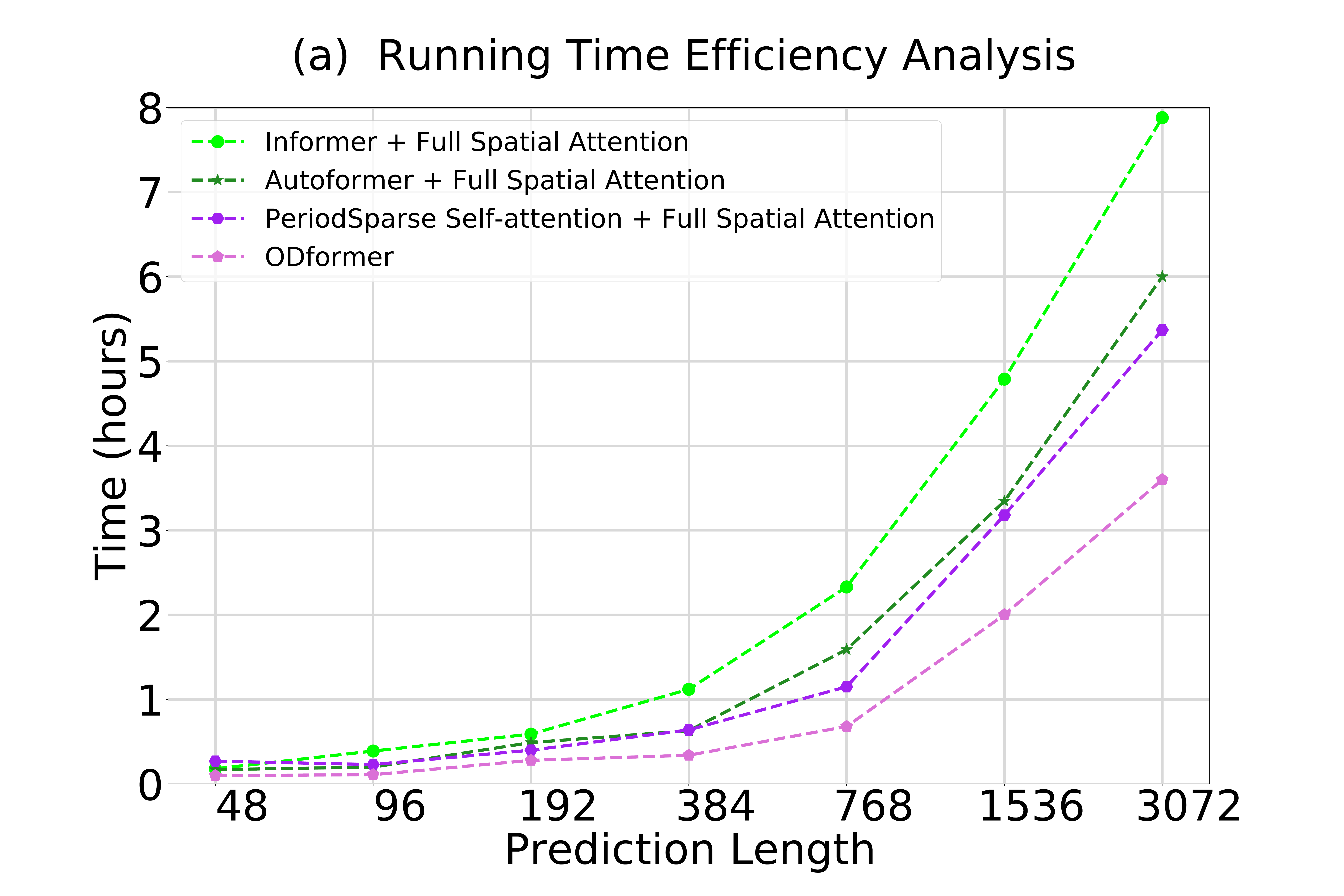}
			\label{fig:7a}
		\end{subfigure}
		\begin{subfigure}{.4\textwidth}
			\centering
			\includegraphics[width=1\textwidth]{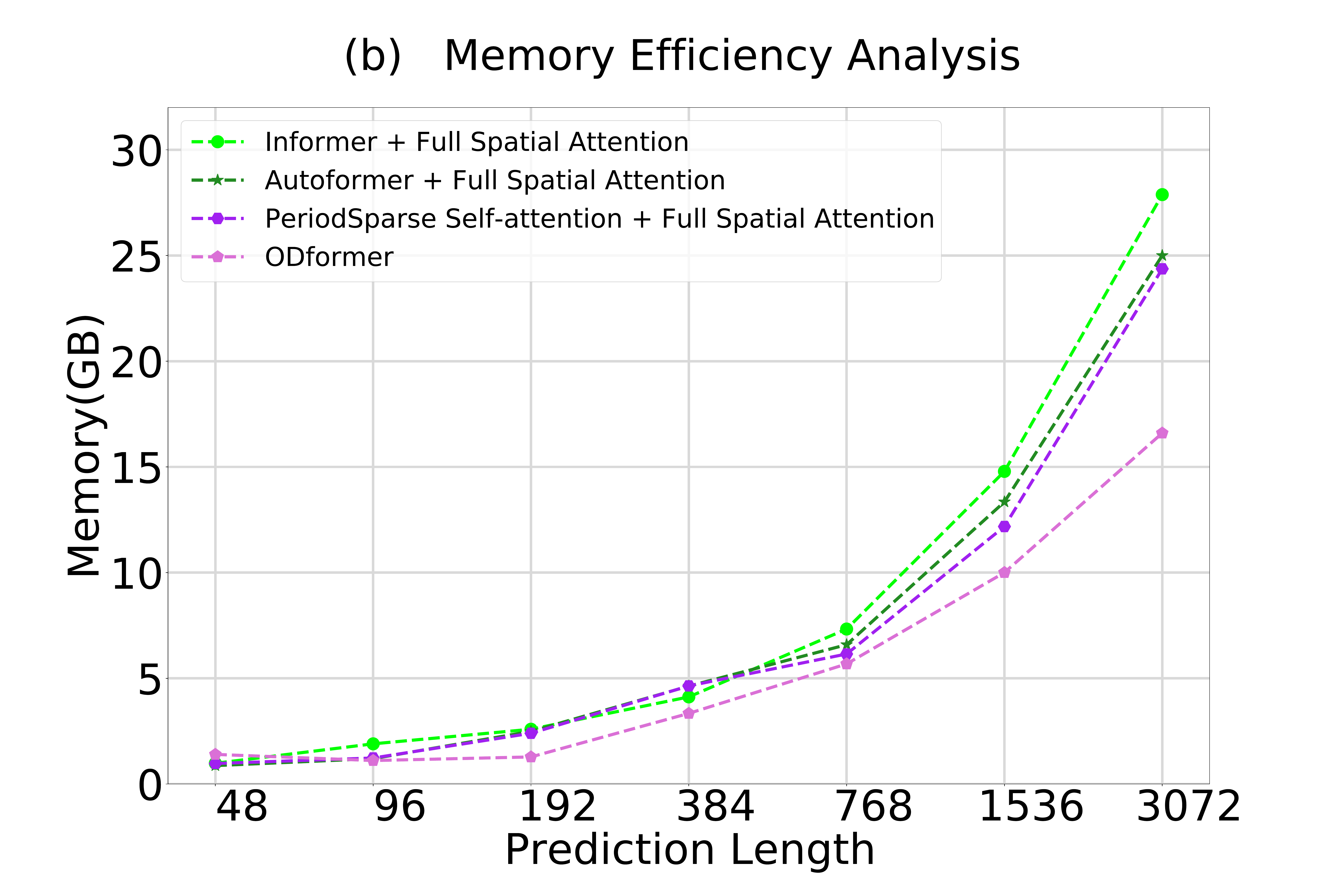}
			\label{fig:7b}
		\end{subfigure}
		\vspace{-0.5cm}
  \caption{Efficiency Analysis. For a fair comparison, we combine Informer and Autoformer with full spatial attention  and replace OD Attention with full spatial attention to verify its efficiency. The output length grows exponentially with the input length 48 on the Abilene dataset.}	\vspace{-.5cm}
  \label{fig:7}
\end{figure}
\subsubsection{Parameter sensitivity: }We perform parameter sensitivity analysis on the spatial dependency and temporal dependency modules of the ODformer model, as shown in Figure \ref{fig:6}.
\\\textit{(1) Spatial share coefficient:} We conduct prediction experiments with the shortest $I-O$ setting on each of the five datasets to analyze the spatial share coefficient $\alpha$. To better represent the trend, we use Min-Max Normalization to process the raw MSE values. We find that the $\alpha$ at the lowest MSE is different on different datasets. On more geographically related datasets, the smaller $\alpha$, the smaller the MSE. Therefore, in the end, we choose the appropriate $\alpha$ according to different application scenarios. When $\alpha$ approaches  extreme values, MSE will rise sharply, which shows the necessity of introducing two different spatial dependencies.
\\\textit{(2) Sparsity control factor:} We set the input sequence length to 192 on the Abilene dataset, and the sparsity control factors were taken from $\{0.25,0.5,1,2,4,8\}$ for experiments. It can be observed that when the control factor is too large, the overall MSE is lower, which proves the necessity of introducing sparse attention for long sequence prediction. In practical applications, we set the control factor to 2.

\subsubsection{Efficiency analysis}
We analyze the time and memory complexity of ODformer here. The complexity in time and memory of ODformer for time-series length L is $\mathcal{O}(L\ln L)$, and the complexity in time and memory of OD Attention on each timestep is $\mathcal{O}(N\ln N)$ (N is the number of regions).  For fairness of comparison, we combine the temporal sparse attention models with  full spatial attention mechanisms. Figure \ref{fig:7} shows the superiority of ODformer in terms of time and memory consumption.

\section{Conclusion}
In this paper, we study the long sequence OD matrix forecasting problem and propose ODformer to forecast long OD matrix series. 
Specifically, we design the OD Attention mechanism and PeriodSparse Self-attention  to deal with the challenges of the variability of spatial-temporal dependencies in various application scenarios and LSTF of OD matrices. Experiments on real-world data in diverse scenarios demonstrate the effectiveness of ODformer in improving the capability of long sequence OD matrix forecasting problems.
\section{Acknowledgments and Disclosure of Funding}
This work was supported by the National Natural Science Foundation of China under Grants 62177015.
%
%



\end{document}